\documentclass{article}
\usepackage[preprint]{neurips_2026}
\usepackage[utf8]{inputenc}
\usepackage[T1]{fontenc}
\usepackage{hyperref}
\usepackage{url}
\usepackage{booktabs}
\usepackage{amsfonts}
\usepackage{amsmath}
\usepackage{nicefrac}
\usepackage{microtype}
\usepackage{xcolor}
\usepackage{graphicx}
\usepackage{subcaption}
\usepackage{multirow}
\usepackage{array}
\usepackage{enumitem}
\usepackage{float}
\usepackage{amsthm}
\usepackage{siunitx}
\newcommand{\creg}{\textsc{CREG}}

\title{CREG: Compass Relational Evidence Graph for Characterizing Directional Structure in VLM Spatial-Reasoning Attribution}

\author{%
\parbox[t]{0.33\textwidth}{\centering\normalfont
\textbf{Kaizhen Tan} \\
Carnegie Mellon University \\
Pittsburgh, PA, USA \\
\texttt{kaizhent@cmu.edu}}%
\parbox[t]{0.33\textwidth}{\centering\normalfont
\textbf{Yang Feng} \\
Columbia University \\
New York, NY, USA \\
\texttt{yf2764@columbia.edu}}%
\parbox[t]{0.33\textwidth}{\centering\normalfont
\textbf{Heqing Du} \\
Columbia University \\
New York, NY, USA \\
\texttt{hd2613@columbia.edu}}%
}

\begin{document}
\maketitle

\begin{abstract}
Standard attribution heatmaps show where a vision-language model (VLM) focuses, but they do not reveal whether the recovered evidence is organized by the queried spatial relation or merely reflects image layout. To address this problem, we introduce \creg{} (\textbf{C}ompass \textbf{R}elational \textbf{E}vidence \textbf{G}raph), a training-free diagnostic framework that converts token-level attribution into a reference-centered compass distribution and measures its directional alignment. CREG provides a shared directional readout across attribution methods and makes comparison with geometric controls explicit. Across three spatial-relation benchmarks, box-only geometry achieves Direction Alignment Error $28.4^\circ$--$34.4^\circ$ lower than the best current model-based attribution method on each dataset, leaving a substantial gap between attribution structure and simple target localization. To examine this gap, we apply a diagnostic battery including target intervention, reference-center randomization, and variance partition. Taken together, the results suggest that the directional structure recoverable from current attribution methods is limited and often mixed with image layout. We further find that higher task accuracy does not reliably coincide with better directional attribution: small-scale LoRA training and newer model generations can improve task accuracy while leaving Direction Alignment Error unchanged or worse. These findings characterize what current attribution methods reveal rather than the model's internal spatial representation. CREG provides a controlled protocol for testing whether improvements in spatial reasoning are accompanied by more directionally organized evidence.
\end{abstract}

\section{Introduction}
\label{sec:intro}

Vision-language models (VLMs) such as Qwen2-VL~\citep{wang2024qwen2vl}, LLaVA~\citep{liu2023visual}, and InternVL~\citep{chen2024internvl} can answer spatial questions such as ``Is the cup to the left of the bottle?'' or ``Where is the chair relative to the table?'' A correct answer, however, does not tell us how the model organizes visual evidence for that decision. Standard attribution methods can highlight which regions are important, but they mainly answer a location question: \emph{where} does the model focus? For spatial reasoning, this is only part of the picture. What is often missing is a directional view: whether the recovered evidence is organized in a way that matches the queried relation between a reference object and a target object.

This distinction matters because spatial attribution is easily confounded by image layout. A saliency map may place strong mass on the target object, yet that pattern does not by itself show that the attribution is organized by the queried direction. It may simply reflect the target's screen position, object size, or other layout factors. In other words, a heatmap can look plausible even when it does not provide clear evidence that the attribution captures relational direction. Existing attribution visualizations make this issue difficult to evaluate, because they show relevance in image coordinates rather than in a relation-centered coordinate system.

To address this problem, we introduce \textbf{CREG} (\textbf{C}ompass \textbf{R}elational \textbf{E}vidence \textbf{G}raph), a training-free diagnostic framework for analyzing spatial attribution in VLMs. CREG takes any token-level attribution signal and projects it into a reference-centered polar coordinate system. The output is a compass distribution that summarizes how attribution mass is organized around the reference object. This gives a shared directional readout across attribution methods and models. Instead of asking only whether attribution overlaps with salient image regions, CREG asks a more specific question: \emph{is the recovered attribution directionally aligned with the spatial relation under query?}

Using this framework, we compare several attribution methods on three spatial-relation benchmarks. A consistent pattern appears across settings: simple box-only geometry is much more directionally aligned than current model-based attribution methods. For the best model-based method on each dataset, the gap is $28.4^\circ$--$34.4^\circ$ in Direction Alignment Error (DAE). This result identifies a stable gap under the shared directional readout, while its source is examined by the diagnostic analyses below. For this reason, we treat CREG as a diagnostic tool rather than as direct evidence about model internals.

We then apply a diagnostic battery to examine what this gap may mean in practice. Our analyses include target intervention, reference-center randomization, and variance partition. Taken together, these results suggest a conservative conclusion: under current attribution methods, the directional structure that CREG can recover is limited and often mixed with image layout. We further study whether better task performance is accompanied by better directional attribution. In our experiments, the answer is not consistently yes. Small-scale LoRA training and newer model generations can improve answer accuracy while leaving DAE unchanged or worse.

Our goal is therefore not to claim that current VLMs lack internal spatial representations. Instead, we aim to provide a controlled framework for measuring what present attribution methods reveal about spatial reasoning behavior. This distinction is important. If task accuracy improves while directional attribution does not, then accuracy alone is not enough to support claims about how spatial reasoning is carried out. A diagnostic framework that makes this gap measurable can help separate progress in task performance from progress in interpretable directional evidence.

Our contributions are as follows:
\begin{enumerate}
    \item We introduce CREG, a training-free and method-agnostic framework that converts token-level attribution into a reference-centered compass distribution, enabling directional evaluation on a shared scale.
    \item We provide a controlled protocol for comparing spatial attribution against geometric baselines, and show a large and consistent gap between current model-based attribution and simple box-based geometry.
    \item We use a diagnostic battery to examine this gap and find converging evidence that the directional structure recoverable from current attribution methods is limited and frequently entangled with layout.
    \item We show that, in our setting, improvements in task accuracy do not reliably co-occur with improvements in directional attribution.
\end{enumerate}

\begin{figure*}[t]
\centering
\includegraphics[width=\linewidth]{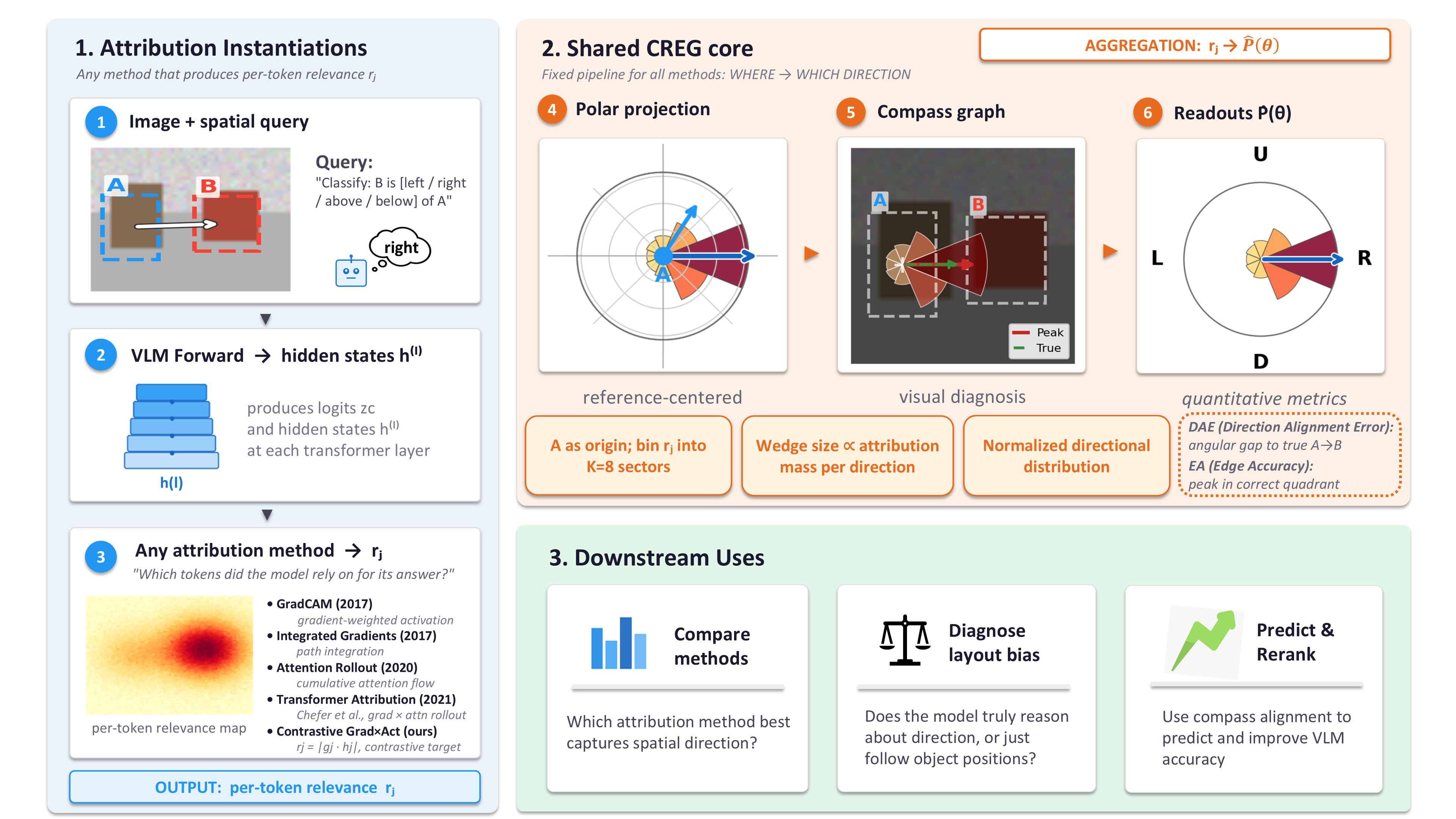}
\caption{\creg{} framework overview. \textbf{Left:} an image-query pair is processed by a VLM, and an attribution method produces per-token relevance scores. \textbf{Upper right:} the shared \creg{} core converts these scores into a reference-centered compass distribution and evaluates directional alignment with DAE and EA. \textbf{Lower right:} the resulting compass readout supports cross-method comparison, comparison against geometric controls, and diagnostic analysis under a shared protocol.}
\label{fig:framework}
\end{figure*}

\section{Related Work}
\label{sec:related}

\paragraph{Spatial reasoning in VLMs.}
Spatial reasoning remains a persistent weakness of vision-language models. Benchmarks such as VSR~\citep{liu2023vsr}, WhatsUp~\citep{kamath2023whatsup}, SpatialSense~\citep{yang2019spatialsense}, and SpatialBot~\citep{cai2024spatialbot} show that strong general VQA performance does not imply reliable understanding of object relations in space. Recent analyses further suggest that part of this difficulty is tied to how VLMs allocate visual attention during spatial tasks~\citep{chen2025adaptvis}. Our work is related to this line of research, but its goal is different. We do not propose a new model for better spatial reasoning accuracy. Instead, we ask how the attribution evidence of existing VLMs is organized, and whether the recovered structure is directionally aligned with the queried relation or mainly reflects image layout.

\paragraph{Attribution and the limits of heatmap explanation.}
A large body of work studies attribution through heatmap-style explanations, including Grad-CAM~\citep{selvaraju2017gradcam}, attention rollout~\citep{abnar2020attention}, Integrated Gradients~\citep{sundararajan2017integrated}, and Transformer Attribution~\citep{chefer2021transformer}. These methods are useful for showing where relevance is concentrated, but for spatial-relation tasks they do not directly answer a more specific question: whether attribution is organized in the direction implied by the query. More broadly, prior work has repeatedly cautioned against treating saliency or attention maps as direct explanations of model reasoning. \citet{adebayo2018sanity} show that some saliency maps remain visually plausible even after model or label randomization, and \citet{jain2019attention} argue that attention weights should not automatically be read as faithful explanations. Our motivation is closely connected to this concern. Rather than assuming that a plausible heatmap already captures relational evidence, \creg{} re-expresses attribution in a reference-centered directional frame and measures how much directional structure can actually be recovered.

\paragraph{Localization, evidence access, and answer correctness.}
Recent work suggests that spatial performance depends partly on whether a model can localize relevant objects and use that information effectively. \citet{ranasinghe2024localize} show that improving object localization can improve downstream spatial reasoning, which indicates that localization is one important ingredient of success on these tasks. At the same time, having access to relevant evidence is not the same as using it correctly. \citet{liu2026seeing} show that VLMs can attend to visually relevant regions and still produce the wrong answer. This distinction is important for our setting. A model may answer correctly without yielding attribution that is clearly organized by the queried direction, and a visually plausible attribution map may still fail to expose relational structure. \creg{} is designed to study this gap by separating directional alignment from answer accuracy, rather than assuming that the two should coincide.

\paragraph{Mechanistic studies of spatial behavior.}
Another related direction examines whether specific internal components of VLMs play specialized roles in spatial reasoning. For example, \citet{ma2026attention} identify sparse attention heads associated with spatial functions and study how ablating or amplifying them affects behavior. This line of work moves closer to causal analysis of internal mechanisms. Our focus is different and more limited. We do not claim to directly recover the model's internal spatial representation. Instead, we study what current attribution methods make visible when spatial reasoning is viewed through a directional readout. In this sense, our framework is complementary to mechanistic analysis. If directional structure exists internally but is only weakly reflected in standard attribution maps, \creg{} provides a way to quantify that mismatch.

\section{Method}
\label{sec:method}

CREG has two components. The first, which is the main contribution of this paper, is a diagnostic framework for measuring directional organization in spatial attribution. It takes an attribution map, projects it into a reference-centered directional space, and evaluates the resulting structure with shared metrics. The second component is a set of attribution instantiations that provide the token-level relevance signal used by the framework. This distinction is important. Our main goal is not to propose a new attribution method, but to provide a common directional readout for analyzing what different attribution methods expose on spatial-relation tasks.

\subsection{Framework}
\label{sec:framework}

\subsubsection{Problem setup}
\label{sec:problem_setup}

Given an image $I$, a reference object $A$, and a target object $B$, we consider a four-way spatial-relation classification setting with labels \textit{left}, \textit{right}, \textit{above}, and \textit{below}. Each sample includes the object names and bounding boxes for $A$ and $B$. A VLM is prompted to predict the relation of $B$ relative to $A$, and produces logits $\{z_c\}$ over the candidate relations at the final token position.

For the same image-query pair, an attribution method produces a relevance score for each visual token, indicating how much that token contributes to the chosen output. Different attribution methods obtain these scores in different ways, for example through gradient propagation, attention flow, or perturbation. CREG treats these scores as input and asks a separate question: once an attribution signal is available, how is its mass organized relative to the queried spatial relation?

\subsubsection{Reference-centered polar projection}
\label{sec:polar_projection}

Standard heatmaps are defined in image coordinates. This makes them useful for showing \emph{where} attribution is concentrated, but less suitable for evaluating whether attribution is organized in the \emph{direction} implied by the query. CREG addresses this by re-expressing attribution in a polar coordinate system centered on the reference object $A$.

Let the visual token grid have size $H_g \times W_g$. Each grid cell $(u,v)$ is associated with pixel coordinates $(x_{uv}, y_{uv})$ and attribution weight $r_{uv}$. We define the polar coordinates of each cell relative to the center of the reference object $A$ as
\begin{align}
\theta_{uv} &= \operatorname{atan2}\!\left(-(y_{uv}-y_A),\; x_{uv}-x_A\right), \notag\\
\rho_{uv}   &= \sqrt{(x_{uv}-x_A)^2 + (y_{uv}-y_A)^2},
\label{eq:polar}
\end{align}
where $0^\circ$ corresponds to the right direction and $90^\circ$ corresponds to the upward direction.

We then divide the angular space into $K$ sectors, with default $K=8$, and aggregate attribution mass within each sector using a Gaussian distance weighting:
\begin{equation}
P(\theta_k) =
\frac{
\sum\limits_{(u,v)\in S_k} r_{uv}\exp\!\left(-\rho_{uv}^2 / 2\sigma^2\right)
}{
\sum\limits_{k'}
\sum\limits_{(u,v)\in S_{k'}} r_{uv}\exp\!\left(-\rho_{uv}^2 / 2\sigma^2\right)
},
\label{eq:compass_dist}
\end{equation}
where $S_k$ is the set of cells whose angle falls into sector $k$, and $\sigma = 0.6 \times 2.0 \times d_{AB}$ with $d_{AB}$ denoting the center distance between $A$ and $B$.

The resulting normalized distribution
\begin{equation}
\hat{P}(\theta) = [P(\theta_1), \ldots, P(\theta_K)]
\end{equation}
is the \textbf{compass attribution distribution}. It summarizes how attribution mass is organized around the reference object in directional terms. This representation does not assume that the reference center is a privileged internal origin of the model. It is a coordinate choice used to test whether the recovered attribution exhibits directional alignment under a relation-centered readout.

\paragraph{Compass visualization.}
To make the directional structure directly interpretable, we visualize $\hat{P}(\theta)$ as an image-overlay compass graph. Each angular sector is drawn as a wedge centered at $A$, with radius proportional to $P(\theta_k)$ and color intensity indicating its magnitude. We also overlay two arrows: one for the peak direction of the compass distribution, and one for the geometric direction from $A$ to $B$. Compared with a standard heatmap, this visualization makes it easier to see whether attribution is directionally aligned, diffuse, or inconsistent with the queried relation.

\begin{figure*}[t]
\centering
\includegraphics[width=\linewidth]{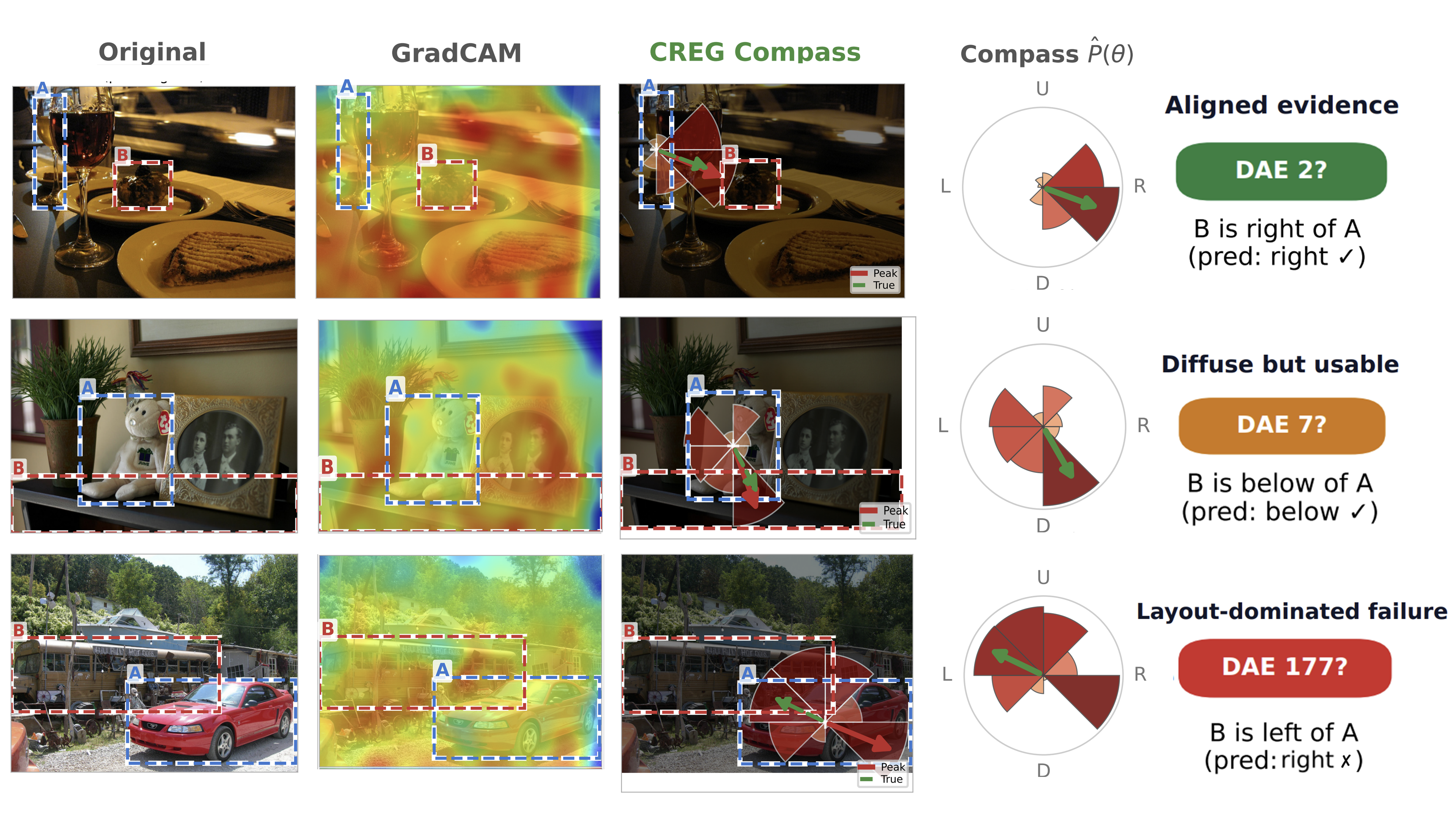}
\caption{From heatmap localization to directional evidence on COCO-Pairs. Each row shows, from left to right, the original image with boxes $A$ and $B$, a GradCAM heatmap, the \creg{} compass overlay, and the compass distribution $\hat{P}(\theta)$. \textbf{Top:} aligned evidence, with the compass peak pointing toward $B$. \textbf{Middle:} diffuse but usable signal, where attribution mass is spread but the peak remains broadly correct. \textbf{Bottom:} a layout-consistent failure under the \creg{} readout, where the compass peak does not align with the queried relation.}
\label{fig:heatmap_to_compass}
\end{figure*}

\subsubsection{Directional metrics}
\label{sec:metrics}

We use two primary metrics throughout the paper.

\paragraph{Direction Alignment Error (DAE).}
DAE measures the circular angular distance between the peak direction of the compass distribution and the geometric direction from $A$ to $B$:
\begin{equation}
\mathrm{DAE} =
\left|
\left(
(\hat{\theta}_{\mathrm{peak}} - \theta_{\mathrm{true}} + 180^\circ) \bmod 360^\circ
\right)
- 180^\circ
\right|,
\label{eq:dae}
\end{equation}
where $\hat{\theta}_{\mathrm{peak}}$ is the center angle of the sector with the largest compass mass, and
\begin{equation}
\theta_{\mathrm{true}} = \operatorname{atan2}\!\left(-(y_B-y_A), \, x_B-x_A\right)
\end{equation}
is the ground-truth geometric direction from the reference object to the target object. DAE ranges from $0^\circ$ to $180^\circ$, and lower values indicate better directional alignment.

\paragraph{Edge Accuracy (EA).}
EA measures whether the peak direction falls into the correct quadrant:
\begin{equation}
\mathrm{EA} = \frac{1}{N}\sum_{i=1}^{N} \mathbf{1}\!\left[\mathrm{DAE}_i \le 45^\circ\right].
\label{eq:ea}
\end{equation}
This metric gives a coarse but intuitive measure of directional correctness. For a random baseline, DAE is expected to be close to $90^\circ$ and EA to be close to $0.25$.

These two metrics serve different roles. DAE captures the magnitude of angular mismatch on a continuous scale, while EA reflects whether the recovered direction is broadly correct at the quadrant level. We use DAE as the main metric and report EA as a supporting summary.

\subsubsection{Protocol summary}
\label{sec:protocol}

The CREG protocol is designed to be reusable across models and attribution methods. For each image-query pair, the input consists of the image, the bounding boxes of the reference and target objects, and any token-level attribution signal $\{r_j\}$. CREG then applies the same reference-centered projection, constructs the same compass distribution, and evaluates the result with the same directional metrics.

This shared protocol makes comparison across attribution methods explicit. It also makes comparison to geometric controls explicit. In particular, we always report three non-model baselines: a geometry oracle, a box-only control, and a random attribution baseline. The geometry oracle gives the ideal direction by construction. The box-only control places uniform attribution mass inside the target box and therefore provides a simple localization-based reference point. The random baseline calibrates chance performance. These controls do not answer the same scientific question, but together they help interpret what a given attribution pattern means under the CREG readout.

\subsection{Attribution instantiations}
\label{sec:instantiations}

The framework above accepts any token-level relevance signal as input. In this sense, CREG is method-agnostic. We evaluate several attribution instantiations in order to compare how different attribution families behave under the same directional measurement protocol. These instantiations are not the main novelty claim of the paper. They are interchangeable signal sources for the shared framework.

\paragraph{Transformer Attribution.}
We implement Transformer Attribution~\citep{chefer2021transformer} as gradient-weighted attention rollout targeted to the relation output. Empirically, this is one of the strongest attribution instantiations in our main comparisons.

\paragraph{Contrastive multi-layer Grad$\times$Act.}
We also evaluate a contrastive multi-layer Grad$\times$Act signal. For transformer layers $\mathcal{L} = \{l_1,\ldots,l_L\}$, the token relevance at layer $l$ is
\begin{equation}
r^{(l)}_j = \left| \sum_d g^{(l)}_{j,d} \cdot h^{(l)}_{j,d} \right|,
\quad
g^{(l)} = \nabla_{h^{(l)}} \tau,
\label{eq:gradxact}
\end{equation}
where $h^{(l)}$ is the hidden state and $\tau$ is a contrastive target defined below. We aggregate layers with softmax weights proportional to their signal magnitude:
\begin{equation}
w_l = \frac{\exp\!\left(\max_j r^{(l)}_j\right)}
{\sum_{l'} \exp\!\left(\max_j r^{(l')}_j\right)},
\quad
r_j = \sum_l w_l \, r^{(l)}_j.
\label{eq:layeragg}
\end{equation}
The contrastive target is
\begin{equation}
\tau = z_{\mathrm{tgt}} - z_{\mathrm{neg}}, \quad
z_{\mathrm{neg}} =
\begin{cases}
z_{\hat{c}}, & \hat{c} \neq \mathrm{tgt} \\
z_{\mathrm{2nd}}, & \text{otherwise}
\end{cases}
\label{eq:contrastive_target}
\end{equation}
where $z_{\mathrm{tgt}}$ is the logit of the target relation, $\hat{c}$ is the predicted class, and $z_{\mathrm{2nd}}$ is the second-largest logit.

\paragraph{Other attribution methods and controls.}
In addition to the two methods above, we evaluate Grad-CAM~\citep{selvaraju2017gradcam}, attention rollout~\citep{abnar2020attention}, Integrated Gradients~\citep{sundararajan2017integrated}, gradient norm, and a perturbation-based baseline using RISE~\citep{petsiuk2018rise}. All of them share the same CREG projection and differ only in the relevance signal they provide. We further compare them with three geometric controls: a geometry oracle, a box-only control, and a random baseline. Full implementation details are given in the appendix.

\section{Experiments}
\label{sec:experiments}

\subsection{Setup}
\label{sec:setup}

\paragraph{Models.}
Our primary model is Qwen2-VL-7B-Instruct~\citep{wang2024qwen2vl}, which we use for the most complete comparison across attribution methods. To test whether the framework transfers across architectures, we further evaluate Qwen2.5-VL-7B, Qwen2-VL-2B, and LLaVA-1.5-7B~\citep{liu2023visual}. Unless otherwise stated, all models are used in inference mode without fine-tuning and run in bfloat16 on a single A800-80GB GPU.

\paragraph{Datasets.}
We evaluate on three spatial-relation benchmarks. COCO-Pairs contains 300 samples constructed from COCO val2017 with matched object bounding boxes. VG-Spatial contains 296 samples from Visual Genome~\citep{krishna2017visual} with object annotations. VSR~\citep{liu2023vsr} provides an additional 240-sample evaluation set with matched boxes and is used as a supplementary benchmark. The main paper focuses on these three datasets because they allow the same directional readout to be computed under explicit object localization.

\paragraph{Attribution methods.}
All attribution methods are evaluated under the same CREG projection. The only difference is the token-level relevance signal supplied to the framework. We compare methods from three attribution families. The gradient-based family includes Transformer Attribution~\citep{chefer2021transformer}, Grad-CAM~\citep{selvaraju2017gradcam}, contrastive multi-layer Grad$\times$Act, Integrated Gradients~\citep{sundararajan2017integrated}, single-layer Grad$\times$Act, and gradient norm. The attention-based family includes attention rollout~\citep{abnar2020attention}. The perturbation-based family includes RISE~\citep{petsiuk2018rise}. We also report three geometric controls: a geometry oracle, a box-only control, and a random baseline.

\paragraph{Evaluation protocol.}
Unless otherwise stated, all methods use the same CREG configuration: $K=8$ compass sectors, the same Gaussian distance weighting, and the same directional metrics (DAE and EA). This shared setup is important because it isolates differences in attribution signal from differences in directional readout. Additional implementation details, robustness checks, and ablations are provided in the appendix.

\subsection{Main results}
\label{sec:main_results}

\begin{table}[t]
\centering
\small
\setlength{\tabcolsep}{2.6pt}
\caption{Framework-level comparison on Qwen2-VL-7B. All methods use the same \creg{} directional readout; only the attribution signal differs. Bold indicates the best model-based score for each metric.}
\label{tab:main}
\begin{tabular}{
@{}l
S[table-format=2.1]
S[table-format=1.3]
S[table-format=2.1]
S[table-format=1.3]
S[table-format=2.1]
S[table-format=1.3]
@{}
}
\toprule
& \multicolumn{2}{c}{\textbf{VSR} (240)} & \multicolumn{2}{c}{\textbf{COCO-Pairs} (300)} & \multicolumn{2}{c}{\textbf{VG-Spatial} (296)} \\
\cmidrule(lr){2-3} \cmidrule(lr){4-5} \cmidrule(lr){6-7}
\textbf{Method} & {DAE$\downarrow$} & {EA$\uparrow$} & {DAE$\downarrow$} & {EA$\uparrow$} & {DAE$\downarrow$} & {EA$\uparrow$} \\
\midrule
Geometry oracle & 0.0 & 1.000 & 0.0 & 1.000 & 0.0 & 1.000 \\
Box-only & 32.7 & 0.733 & 15.0 & 0.973 & 14.2 & 0.963 \\
\midrule
Transformer Attribution & {\bfseries 61.1} & {\bfseries 0.492} & {\bfseries 49.4} & {\bfseries 0.620} & 49.2 & 0.679 \\
GradCAM & 63.1 & 0.471 & 53.5 & 0.547 & 54.8 & 0.578 \\
RISE (perturbation)$^\ddagger$ & 62.0 & 0.479 & 56.9 & 0.540 & {\bfseries 47.1} & 0.622 \\
Contrastive Grad$\times$Act & 71.0 & 0.417 & 66.3 & 0.470 & 52.9 & 0.639 \\
\midrule
Attention rollout & 74.8 & 0.379 & 76.1 & 0.367 & 72.0 & 0.419 \\
Integrated Gradients$^\dagger$ & 75.3 & 0.400 & 79.1 & 0.350 & 90.7 & 0.333 \\
Random & 89.9 & 0.221 & 94.2 & 0.257 & 91.4 & 0.250 \\
\bottomrule
\end{tabular}

\vspace{3pt}
\parbox{\linewidth}{\scriptsize $^\dagger$ IG (50-step) evaluated on 60-sample subsets per dataset. $^\ddagger$ RISE uses 150 random masks per sample and is evaluated on the full dataset.}
\end{table}

Table~\ref{tab:main} presents the main comparison on Qwen2-VL-7B. A clear pattern appears across all three datasets. The geometry oracle gives DAE $= 0^\circ$ by construction. The random baseline stays close to chance level. Between these extremes, the box-only control is substantially more directionally aligned than all model-based attribution methods. Its DAE ranges from $14.2^\circ$ to $32.7^\circ$, whereas the best model-based methods remain in the range of $47.1^\circ$ to $61.1^\circ$. The exact best-method gaps to box-only are $28.4^\circ$ on VSR, $34.4^\circ$ on COCO-Pairs, and $32.9^\circ$ on VG-Spatial.

Among model-based methods, Transformer Attribution is strongest on VSR and COCO-Pairs, while RISE gives the lowest DAE on VG-Spatial. The second-best method also varies by dataset: RISE is second on VSR, Grad-CAM is second on COCO-Pairs, and Transformer Attribution is second on VG-Spatial. By contrast, attention rollout and Integrated Gradients are consistently weaker under the same directional readout. These comparisons support one stable empirical conclusion: under the CREG measurement protocol, current attribution methods recover substantially weaker directional alignment than a simple target-box localization control.

Table~\ref{tab:main} also gives the method-level differences directly. Transformer Attribution improves over Contrastive Grad$\times$Act by $9.9^\circ$ on VSR, $16.9^\circ$ on COCO-Pairs, and $3.7^\circ$ on VG-Spatial, for a sample-count-weighted mean difference of approximately $10.2^\circ$. More importantly, model-based attribution remains substantially worse than the box-only control; for the best model-based method on each dataset, the DAE gaps are $28.4^\circ$, $34.4^\circ$, and $32.9^\circ$ ($p<.001$ for the main comparisons). We therefore interpret the box-only gap as a stable empirical property of the current readout, while reserving its causal interpretation for the diagnostic analyses in Section~\ref{sec:diagnostic_case}.

This result should be interpreted carefully. The box-only control is not a neutral baseline for spatial reasoning. By construction, it concentrates mass inside the target object and therefore aligns well with a direction metric defined from the reference object to the target object. For this reason, the comparison does not imply that box-only geometry is ``doing reasoning.'' The more appropriate interpretation is narrower: current model-based attribution is considerably more diffuse, less directionally organized, or both, than simple target localization under the same readout. Whether this gap reflects layout effects, attribution-method limitations, or a combination of the two is examined in Section~\ref{sec:diagnostic_case}. Additional sanity checks on synthetic attribution maps are reported in Appendix~\ref{app:synthetic_sanity}.

Overall, the main result of this section is not that attribution has been fully explained, but that the shared directional readout reveals a robust mismatch between current attribution maps and simple geometric localization. This mismatch is the central empirical motivation for the diagnostic analyses that follow. Cross-model results further show that the strongest attribution plug-in is model-dependent under the same directional readout; we defer the full comparison to Appendix~\ref{app:cross_model}.

\begin{figure}[t]
\centering
\includegraphics[width=0.98\linewidth]{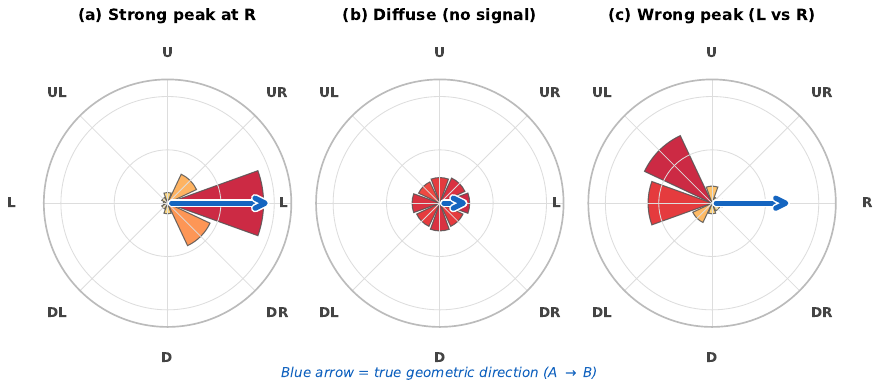}
\caption{Examples of compass distributions from \creg{} under the shared directional readout. Blue arrow denotes the true geometric direction. (a) Strong alignment. (b) Diffuse signal. (c) Incorrect peak.}
\label{fig:compass_examples}
\end{figure}

Qualitatively, the compass readout often falls into three recurring patterns: \emph{strong alignment}, where the peak direction points toward the target; \emph{diffuse signal}, where attribution mass is spread and only weakly directional; and \emph{incorrect peak}, where the dominant direction does not match the queried relation. These examples are illustrated in Figure~\ref{fig:compass_examples}.

\section{Diagnostic Case Study}
\label{sec:diagnostic_case}

Section~\ref{sec:main_results} shows a clear framework-level gap between current model-based attribution and simple box-based geometry. A natural next question is how this gap should be interpreted. CREG by itself does not resolve this question. It provides a directional readout, not a direct probe of model internals. We therefore apply three diagnostics---target intervention, reference-center randomization, and variance partition---to test whether the recovered attribution behaves as if it were strongly conditioned on relational direction.

\paragraph{Target intervention.}
Our first diagnostic asks whether the compass distribution changes when the probed relation label changes while the image is fixed. In practice, the response is weak: target-logit intervention yields PTA $=24.9\%$, essentially at the random baseline of $25\%$, and the Jensen--Shannon divergence between opposite-probe compass distributions is only $0.004$. A stronger text-query role-swap test produces mean peak shift $98.5^\circ \pm 52.7^\circ$, suggesting that some query-sensitive component exists, but it is not dominant under target-logit intervention alone. This result shows that the compass readout changes only weakly under target-logit intervention.

\paragraph{Reference-center randomization.}
Our second diagnostic tests whether the true reference object provides a privileged origin for the compass projection. Re-projecting the same attribution maps from alternative centers yields no advantage for the true center: one-sided permutation tests give $p \ge 0.93$ across conditions, and random centers are statistically indistinguishable from the true center. Background centers can even produce lower DAE. We exclude the wrong-object center-at-$B$ condition from interpretation because the direction from that center to $B$ is a zero vector. This pattern indicates that the recovered compass distributions do not show a robust preference for the true reference object as the projection center.

\paragraph{Variance partition.}
Our third diagnostic estimates how much of the within-image variation in the compass readout is explained by direction change versus layout change. In the controlled Fix-B-Move-A setting, direction change explains only a small portion of within-image variance ($\eta^2=0.013$), while location change explains a similarly small but somewhat larger portion ($\eta^2=0.069$). Overall variance remains dominated by image identity. This suggests that the directional component recoverable by the CREG readout is modest relative to broader layout- and image-dependent effects.

\paragraph{Summary.}
The three diagnostics are consistent with one another and support the same qualitative interpretation. Target intervention is near-random, though answer-token gradient correlation may weaken the test. Reference-center randomization shows no advantage for the true center, though target-concentrated attribution can also produce this pattern. Variance partition finds only a small recoverable directional component once image identity is controlled for. Taken together, these results support a conservative interpretation: under current attribution methods, the directional structure recoverable by CREG is limited and often mixed with image layout. The signal-injection positive control in Appendix~\ref{app:signal_injection} confirms that the CREG readout can recover directional signal when such signal is sufficiently present in the attribution map.

\section{Accuracy Gains Without Reliable Directional Improvement}
\label{sec:accuracy_without_direction}

A related question is whether better task performance leads to more directionally organized attribution. In our experiments, the answer is not consistently yes. Across controlled LoRA training and cross-generational comparison, answer accuracy can improve while Direction Alignment Error (DAE) remains unchanged or becomes worse.

\paragraph{Controlled training and model comparison.}
We first apply small-scale LoRA training to Qwen2-VL-7B under three strategies that share the same data, learning rate, and number of epochs: direction-supervised training, target-localization training, and relation-contrastive training. All three improve task accuracy relative to the corresponding intervention baseline, but none yields a stable DAE improvement across datasets. We then compare successive model generations using each model's strongest attribution method under the shared CREG protocol. The same pattern remains: newer models can achieve higher task accuracy without lower DAE. Full numbers for the LoRA intervention and the cross-generational comparison are reported in Appendix~\ref{app:intervention_full}.

\paragraph{Scope of the claim.}
We do not claim that task accuracy and directional attribution are fundamentally unrelated. Our claim is narrower: in our setting, they do not reliably improve together. The LoRA intervention is small in scale, the VSR evaluation uses the subset with matched boxes, and the model comparisons are based on publicly available checkpoints. Under these conditions, benchmark accuracy should not be treated as a proxy for directionally organized attribution.

\paragraph{Connection to the main argument.}
Taken together with the earlier sections, these results reinforce the central message of the paper. CREG exposes a persistent gap between answer correctness and directional attribution quality under a shared readout. This gap appears not only across attribution methods, but also across training interventions and model variants. For this reason, claims about improved spatial reasoning should be interpreted carefully when they are based only on task accuracy or visually plausible heatmaps.

\section{Conclusion}
\label{sec:conclusion}

We introduced CREG, a training-free diagnostic framework that converts token-level attribution into a reference-centered compass distribution and evaluates its directional alignment on a shared scale. Across models, attribution methods, and spatial-relation benchmarks, CREG reveals a large and consistent gap between current model-based attribution and a box-only geometric control.

Our diagnostic analyses support a narrow interpretation of this gap. Under current attribution methods, the directional structure recoverable by CREG is limited and often mixed with image layout. At the same time, CREG is a readout of attribution behavior rather than a direct probe of internal mechanism. We also find that, in our setting, higher task accuracy does not reliably coincide with better directional attribution.

We view CREG as a measurement tool rather than a final explanation of spatial reasoning in VLMs. Its main value is to make an important distinction measurable: a model can answer spatial questions correctly while current attribution methods still fail to expose clearly directionally organized evidence. Future work can test whether larger-scale training, stronger spatial supervision, different explanation families, or more mechanistic analyses reduce this gap.


\bibliographystyle{plainnat}
\bibliography{references}

\begin{thebibliography}{20}
\providecommand{\natexlab}[1]{#1}
\providecommand{\url}[1]{\texttt{#1}}
\expandafter\ifx\csname urlstyle\endcsname\relax
  \providecommand{\doi}[1]{doi: #1}\else
  \providecommand{\doi}{doi: \begingroup \urlstyle{rm}\Url}\fi

\bibitem[Abnar and Zuidema(2020)]{abnar2020attention}
Samira Abnar and Willem Zuidema.
\newblock Quantifying attention flow in transformers.
\newblock In \emph{Proceedings of the 58th Annual Meeting of the Association
  for Computational Linguistics}, 2020.

\bibitem[Adebayo et~al.(2018)Adebayo, Gilmer, Muelly, Goodfellow, Hardt, and
  Kim]{adebayo2018sanity}
Julius Adebayo, Justin Gilmer, Michael Muelly, Ian Goodfellow, Moritz Hardt,
  and Been Kim.
\newblock Sanity checks for saliency maps.
\newblock In \emph{Advances in Neural Information Processing Systems}, 2018.

\bibitem[Cai et~al.(2024)Cai, Ponomarenko, Yuan, Li, Yang, Dong, and
  Zhao]{cai2024spatialbot}
Wenxiao Cai, Yaroslav Ponomarenko, Jianhao Yuan, Xiaoqi Li, Wankou Yang, Hao
  Dong, and Bo~Zhao.
\newblock {SpatialBot}: Precise spatial understanding with vision language
  models, 2024.

\bibitem[Chefer et~al.(2021)Chefer, Gur, and Wolf]{chefer2021transformer}
Hila Chefer, Shir Gur, and Lior Wolf.
\newblock Transformer interpretability beyond attention visualization.
\newblock In \emph{Proceedings of the IEEE/CVF Conference on Computer Vision
  and Pattern Recognition}, 2021.

\bibitem[Chen et~al.(2025)Chen, Zhu, Zhou, Zhang, Gao, Niebles, Geva, He, Wu,
  and Li]{chen2025adaptvis}
Shiqi Chen, Tongyao Zhu, Ruochen Zhou, Jinghan Zhang, Siyang Gao, Juan~Carlos
  Niebles, Mor Geva, Junxian He, Jiajun Wu, and Manling Li.
\newblock Why is spatial reasoning hard for {VLMs}? an attention mechanism
  perspective on focus areas.
\newblock In \emph{Proceedings of the 42nd International Conference on Machine
  Learning}, pages 9910--9932, 2025.

\bibitem[Chen et~al.(2024)Chen, Wu, Wang, Su, Chen, Xing, Zhong, Zhang, Zhu,
  Lu, Li, Luo, Lu, Qiao, and Dai]{chen2024internvl}
Zhe Chen, Jiannan Wu, Wenhai Wang, Weijie Su, Guo Chen, Sen Xing, Muyan Zhong,
  Qinglong Zhang, Xizhou Zhu, Lewei Lu, Bin Li, Ping Luo, Tong Lu, Yu~Qiao, and
  Jifeng Dai.
\newblock {InternVL}: Scaling up vision foundation models and aligning for
  generic visual-linguistic tasks.
\newblock In \emph{Proceedings of the IEEE/CVF Conference on Computer Vision
  and Pattern Recognition}, pages 24185--24198, 2024.

\bibitem[Hu et~al.(2022)Hu, Shen, Wallis, Allen-Zhu, Li, Wang, Wang, and
  Chen]{hu2022lora}
Edward~J. Hu, Yelong Shen, Phillip Wallis, Zeyuan Allen-Zhu, Yuanzhi Li, Shean
  Wang, Lu~Wang, and Weizhu Chen.
\newblock {LoRA}: Low-rank adaptation of large language models.
\newblock In \emph{International Conference on Learning Representations}, 2022.

\bibitem[Jain and Wallace(2019)]{jain2019attention}
Sarthak Jain and Byron~C. Wallace.
\newblock Attention is not explanation.
\newblock In \emph{Proceedings of the 2019 Conference of the North American
  Chapter of the Association for Computational Linguistics: Human Language
  Technologies}, pages 3543--3556, 2019.

\bibitem[Kamath et~al.(2023)Kamath, Hessel, and Chang]{kamath2023whatsup}
Amita Kamath, Jack Hessel, and Kai-Wei Chang.
\newblock What's ``up'' with vision-language models? investigating their
  struggle with spatial reasoning.
\newblock In \emph{Proceedings of the 2023 Conference on Empirical Methods in
  Natural Language Processing}, 2023.

\bibitem[Krishna et~al.(2017)Krishna, Zhu, Groth, Johnson, Hata, Kravitz, Chen,
  Kalantidis, Li, Shamma, Bernstein, and Fei-Fei]{krishna2017visual}
Ranjay Krishna, Yuke Zhu, Oliver Groth, Justin Johnson, Kenji Hata, Joshua
  Kravitz, Stephanie Chen, Yannis Kalantidis, Li-Jia Li, David~A. Shamma,
  Michael~S. Bernstein, and Li~Fei-Fei.
\newblock Visual genome: Connecting language and vision using crowdsourced
  dense image annotations.
\newblock \emph{International Journal of Computer Vision}, 123:\penalty0
  32--73, 2017.

\bibitem[Liu et~al.(2023{\natexlab{a}})Liu, Emerson, and Collier]{liu2023vsr}
Fangyu Liu, Guy Emerson, and Nigel Collier.
\newblock Visual spatial reasoning.
\newblock \emph{Transactions of the Association for Computational Linguistics},
  11:\penalty0 635--651, 2023{\natexlab{a}}.

\bibitem[Liu et~al.(2023{\natexlab{b}})Liu, Li, Wu, and Lee]{liu2023visual}
Haotian Liu, Chunyuan Li, Qingyang Wu, and Yong~Jae Lee.
\newblock Visual instruction tuning.
\newblock In \emph{Advances in Neural Information Processing Systems},
  2023{\natexlab{b}}.

\bibitem[Liu et~al.(2026)Liu, Chen, Liu, Luo, Tang, Wang, Zeng, Dai, Shi, Wei,
  Lu, Dumoulin, and Tong]{liu2026seeing}
Zhining Liu, Ziyi Chen, Hui Liu, Chen Luo, Xianfeng Tang, Suhang Wang, Jingying
  Zeng, Zhenwei Dai, Zhan Shi, Tianxin Wei, Hanqing Lu, Benoit Dumoulin, and
  Hanghang Tong.
\newblock Seeing but not believing: Probing the disconnect between visual
  attention and answer correctness in {VLMs}.
\newblock In \emph{International Conference on Learning Representations}, 2026.

\bibitem[Ma et~al.(2026)Ma, Yang, Jiang, Liu, Liu, Ao, Ma, Erfani, and
  Bailey]{ma2026attention}
Xueqi Ma, Shuo Yang, Yanbei Jiang, Shu Liu, Zhenzhen Liu, Jiayang Ao, Xingjun
  Ma, Sarah~Monazam Erfani, and James Bailey.
\newblock Attention in space: Functional roles of {VLM} heads for spatial
  reasoning, 2026.

\bibitem[Petsiuk et~al.(2018)Petsiuk, Das, and Saenko]{petsiuk2018rise}
Vitali Petsiuk, Abir Das, and Kate Saenko.
\newblock {RISE}: Randomized input sampling for explanation of black-box
  models.
\newblock In \emph{Proceedings of the British Machine Vision Conference}, 2018.

\bibitem[Ranasinghe et~al.(2024)Ranasinghe, Shukla, Poursaeed, Ryoo, and
  Lin]{ranasinghe2024localize}
Kanchana Ranasinghe, Satya~Narayan Shukla, Omid Poursaeed, Michael~S. Ryoo, and
  Tsung-Yu Lin.
\newblock Learning to localize objects improves spatial reasoning in
  visual-{LLM}s.
\newblock In \emph{Proceedings of the IEEE/CVF Conference on Computer Vision
  and Pattern Recognition}, 2024.

\bibitem[Selvaraju et~al.(2017)Selvaraju, Cogswell, Das, Vedantam, Parikh, and
  Batra]{selvaraju2017gradcam}
Ramprasaath~R. Selvaraju, Michael Cogswell, Abhishek Das, Ramakrishna Vedantam,
  Devi Parikh, and Dhruv Batra.
\newblock {Grad-CAM}: Visual explanations from deep networks via gradient-based
  localization.
\newblock In \emph{Proceedings of the IEEE International Conference on Computer
  Vision}, 2017.

\bibitem[Sundararajan et~al.(2017)Sundararajan, Taly, and
  Yan]{sundararajan2017integrated}
Mukund Sundararajan, Ankur Taly, and Qiqi Yan.
\newblock Axiomatic attribution for deep networks.
\newblock In \emph{Proceedings of the 34th International Conference on Machine
  Learning}, 2017.

\bibitem[Wang et~al.(2024)Wang, Bai, Tan, Wang, Fan, Bai, Chen, Liu, Wang, Ge,
  Fan, Dang, Du, Ren, Men, Liu, Zhou, Zhou, and Lin]{wang2024qwen2vl}
Peng Wang, Shuai Bai, Sinan Tan, Shijie Wang, Zhihao Fan, Jinze Bai, Keqin
  Chen, Xuejing Liu, Jialin Wang, Wenbin Ge, Yang Fan, Kai Dang, Mengfei Du,
  Xuancheng Ren, Rui Men, Dayiheng Liu, Chang Zhou, Jingren Zhou, and Junyang
  Lin.
\newblock {Qwen2-VL}: Enhancing vision-language model's perception of the world
  at any resolution, 2024.

\bibitem[Yang et~al.(2019)Yang, Russakovsky, and Deng]{yang2019spatialsense}
Kaiyu Yang, Olga Russakovsky, and Jia Deng.
\newblock {SpatialSense}: An adversarially crowdsourced benchmark for spatial
  relation recognition.
\newblock In \emph{Proceedings of the IEEE/CVF International Conference on
  Computer Vision}, pages 2051--2060, 2019.

\end{thebibliography}

\newpage
\appendix
\section{Additional Details}
\label{app:additional_details}

This appendix reports only the supplementary analyses that are directly needed to support the main claims of the paper. We keep five components: (i) metric definitions and implementation details, (ii) dataset and annotation details, (iii) robustness of the CREG readout, (iv) additional details for the main diagnostic analyses, and (v) training-intervention details. We omit secondary analyses that do not materially change the main conclusions, including auxiliary faithfulness composites, selective prediction, octant evaluation, failure taxonomies, re-ranking analysis, and other exploratory results.

\section{Metrics and Implementation}
\label{app:metrics_impl}

\subsection{Primary Metrics}
\label{app:primary_metrics}

We use two primary metrics throughout the paper.

\paragraph{Direction Alignment Error (DAE).}
DAE measures the circular angular distance between the peak direction of the compass distribution and the geometric direction from the reference object $A$ to the target object $B$:
\begin{equation}
\mathrm{DAE} =
\left|
\left(
(\hat{\theta}_{\mathrm{peak}} - \theta_{\mathrm{true}} + 180^\circ) \bmod 360^\circ
\right)
- 180^\circ
\right|,
\end{equation}
where $\hat{\theta}_{\mathrm{peak}}$ is the center angle of the sector with maximum compass mass, and
\begin{equation}
\theta_{\mathrm{true}} = \operatorname{atan2}\!\left(-(y_B-y_A), \, x_B-x_A\right).
\end{equation}
DAE lies in $[0^\circ, 180^\circ]$, and lower values indicate better directional alignment.

\paragraph{Edge Accuracy (EA).}
EA measures whether the peak compass direction falls within the correct quadrant:
\begin{equation}
\mathrm{EA} = \frac{1}{N}\sum_{i=1}^{N} \mathbf{1}\!\left[\mathrm{DAE}_i \le 45^\circ\right].
\end{equation}
For a random baseline, DAE is expected to be close to $90^\circ$ and EA to be close to $0.25$.

We use DAE as the primary metric and report EA as a supporting summary. This choice keeps the evaluation focused on directional alignment rather than on a larger collection of secondary scores.

\subsection{Direction Estimator}
\label{app:direction_estimator}

The main paper uses the peak sector of the compass distribution as the directional estimate. To verify that the conclusions do not depend on this choice alone, we also compute a circular-mean direction:
\begin{equation}
\hat{\theta}_{\mathrm{cm}} =
\operatorname{atan2}
\left(
\sum_k P(\theta_k)\sin\theta_k,\,
\sum_k P(\theta_k)\cos\theta_k
\right).
\end{equation}

As a distribution-level alternative, we compute circular-mean DAE from the compass distribution. On 200 COCO-Pairs samples, circular-mean DAE is $49.5^\circ$ compared with peak-based DAE of $56.0^\circ$, with Pearson correlation $r=0.68$. EA shows the same qualitative pattern. The method ranking and the gap to the box-only control remain unchanged. We therefore use peak-based DAE in the main paper for consistency with the discrete sector structure of $\hat{P}(\theta)$.

More broadly, the same qualitative conclusion also holds across the main design choices of the CREG readout. In the original robustness sweep on COCO-Pairs, all four tested models degrade substantially at $\sigma_r=0.3$, while $\sigma_r \in [0.6, 2.0]$ remains stable within about $3^\circ$ DAE and 0.02 EA, with model ranking preserved. Likewise, $K=4$ is consistently too coarse, whereas $K=8$ gives the best DAE/EA balance and preserves the same qualitative method ordering. We therefore do not claim that every possible estimator yields identical numeric values; rather, our claim is that the main method ranking and the gap to simple localization remain stable under reasonable alternatives already tested in the original appendix.

\subsection{Prompt, Hyperparameters, and Attribution Setup}
\label{app:implementation}

Unless otherwise stated, all experiments use the following prompt:
\begin{quote}
\small
\texttt{In this image, where is the \{tgt\} relative to the \{ref\}? Choose one option: 1) to the left, 2) to the right, 3) above, 4) below. Answer with just the number.}
\end{quote}

The default CREG configuration is:
\begin{itemize}
    \item Compass sectors: $K=8$
    \item Direction estimator: peak sector
    \item Gaussian weighting: $\sigma = 0.6 \times 2.0 \times d_{AB}$
    \item Hidden-state layers for multi-layer Grad$\times$Act: $\{-2,-3,-4,-5\}$
    \item Precision: bfloat16
    \item Hardware: single NVIDIA A800-80GB GPU
\end{itemize}

All attribution methods use the same CREG projection and differ only in the token-level relevance signal:
\begin{itemize}
    \item \textbf{Grad-CAM} extracts a gradient-weighted activation map from the last layer of the vision encoder and upsamples it to the visual token grid.
    \item \textbf{Attention rollout} multiplies attention matrices cumulatively across layers with identity residuals, then reads the row corresponding to the last text token over visual-token columns.
    \item \textbf{Transformer Attribution} uses gradient-weighted attention rollout following~\citet{chefer2021transformer}.
    \item \textbf{Contrastive multi-layer Grad$\times$Act} uses the formulation in Section~\ref{sec:instantiations} of the main paper.
    \item \textbf{Integrated Gradients} is evaluated with 50 interpolation steps.
    \item \textbf{RISE} uses 150 random masks per sample.
\end{itemize}

We also report three non-model controls:
\begin{itemize}
    \item \textbf{Geometry oracle:} uses the true box centers directly, giving DAE $=0^\circ$ by construction.
    \item \textbf{Box-only control:} assigns uniform relevance to tokens inside the target box and zero elsewhere, then applies the same CREG projection.
    \item \textbf{Random baseline:} samples token relevance i.i.d.\ from $\mathrm{Uniform}(0,1)$ and applies the same projection.
\end{itemize}

The CREG projection itself adds negligible overhead compared with attribution computation. In our implementation, the end-to-end runtime is dominated by the forward/backward passes required by the attribution method rather than by the compass construction.

\section{Dataset and Annotation Details}
\label{app:data_details}

\subsection{Evaluation Sets}
\label{app:datasets}

We evaluate on three spatial-relation datasets:
\begin{itemize}
    \item \textbf{COCO-Pairs} ($n=300$), constructed from COCO val2017 with matched object bounding boxes.
    \item \textbf{VG-Spatial} ($n=296$), constructed from Visual Genome with object annotations.
    \item \textbf{VSR} ($n=240$), used as a supplementary benchmark after matching entities to COCO object categories with valid boxes.
\end{itemize}

The main paper focuses on these three datasets because all of them allow the same reference-centered directional readout to be computed with explicit object localization.

\subsection{VSR Bounding-Box Matching Audit}
\label{app:vsr_audit}

VSR does not provide native object bounding boxes. We therefore match subject and object names in VSR captions to COCO instance categories using a conservative exact-or-substring matching rule. This produces a smaller but high-confidence evaluation subset. Table~\ref{tab:vsr_audit} reports the coverage.

\begin{table}[t]
\centering
\caption{VSR bounding-box matching audit. Coverage is the fraction of positive VSR samples where both entities are matched to COCO categories. Validity is the fraction of matched samples with non-degenerate boxes.}
\label{tab:vsr_audit}
\small
\begin{tabular}{lccc}
\toprule
Split & Positive & Matched & Coverage / Validity \\
\midrule
Test  & 1181 & 240 & 20.3\% / 100\% \\
Dev   & 564  & 121 & 21.5\% / 100\% \\
Train & 3876 & 828 & 21.4\% / 100\% \\
\midrule
Total & 5621 & 1189 & 21.2\% / 100\% \\
\bottomrule
\end{tabular}
\end{table}

This subset spans all four cardinal relations and is used only as supplementary evidence. COCO-Pairs and VG-Spatial remain the primary evaluation datasets in the main paper.

\section{Robustness of the CREG Readout}
\label{app:robustness}

\subsection{Sensitivity to Compass Design Choices}
\label{app:design_choices}

We test whether the main conclusions depend strongly on the specific CREG hyperparameters.

\paragraph{Distance weighting ($\sigma$).}
Across four models on COCO-Pairs, $\sigma_r = 0.3$ consistently degrades performance, while $\sigma_r \in [0.6, 2.0]$ yields stable results. In this range, DAE typically varies by less than $3^\circ$ and EA by less than 0.02. The model ranking is preserved throughout.

\paragraph{Sector count ($K$).}
We compare $K \in \{4, 8, 16\}$. Using only four sectors is too coarse and weakens both DAE and EA. Increasing beyond $K=8$ does not change the qualitative conclusions. In particular, the method ranking and the gap to the box-only control remain stable. We therefore use $K=8$ as the default setting.

Together, these results indicate that the main findings are not artifacts of one narrow hyperparameter choice.

\subsection{Robustness to Bounding-Box Noise}
\label{app:bbox_noise}

To test sensitivity to imperfect object localization, we add Gaussian noise to both reference and target box centers and re-project the same attribution maps. Across four models on COCO-Pairs, adding 50-pixel noise changes model-based DAE by at most $3.7^\circ$, while the geometry oracle degrades by $24.2^\circ$ at the same noise level. The model ranking remains unchanged. This indicates that CREG does not depend critically on precise box localization.

\section{Additional Diagnostic Details}
\label{app:diagnostic_details}

This section provides concise supporting details for the three diagnostic analyses in Section~\ref{sec:diagnostic_case} of the main paper, together with one positive control showing that the CREG readout can recover directional signal when such signal is present in the attribution map.

\subsection{Target Intervention}
\label{app:target_intervention}

On Qwen2-VL-7B, target-logit intervention over the four candidate relations yields PTA = $24.9\%$, essentially matching the random baseline of $25\%$. The Jensen--Shannon divergence between opposite-probe compass distributions is only $0.004$, indicating that the readout changes little when only the probed label is changed on a fixed image. A stronger text-query role-swap test produces mean peak shift $98.5^\circ \pm 52.7^\circ$, which suggests that some query-sensitive component exists, but it is not dominant under target-logit intervention alone.

\subsection{Reference-Center Randomization}
\label{app:rcr}

We re-project each attribution map from multiple alternative centers: random image locations and sampled background locations. We do not interpret the center-at-$B$ wrong-object condition because the direction from that center to target $B$ is a zero vector. Table~\ref{tab:rcr_full} reports the full result.

\begin{table}[t]
\centering
\caption{Reference-center randomization. $\Delta$ is DAE(alternative) $-$ DAE(true center). Negative $\Delta$ means the alternative center gives lower DAE than the true reference center.}
\label{tab:rcr_full}
\small
\begin{tabular}{llcc}
\toprule
Dataset & Center & DAE & $\Delta$ vs.\ true \\
\midrule
\multirow{3}{*}{COCO-Pairs}
& True reference (A) & 58.1 & --- \\
& Random            & 54.0 & $-4.1$ \\
& Background        & 49.0 & $-9.1$ \\
\midrule
\multirow{3}{*}{VG-Spatial}
& True reference (A) & 54.4 & --- \\
& Random            & 50.4 & $-4.0$ \\
& Background        & 49.6 & $-4.8$ \\
\bottomrule
\end{tabular}
\end{table}

One-sided permutation tests of the hypothesis that the true center gives lower DAE yield $p \ge 0.93$ across conditions. Two-sided tests likewise show no advantage for the true center over random centers. The main conclusion is clear: under current attribution methods, the recovered compass distributions do not show a robust preference for the true reference center.

\subsection{Variance Partition}
\label{app:variance_partition}

We estimate how much of the within-image variance in the compass readout is explained by direction change versus location change using a controlled Fix-B-Move-A setup. The result is summarized in Table~\ref{tab:variance_partition}.

\begin{table}[t]
\centering
\caption{Variance partition in the controlled Fix-B-Move-A setting. Relative share is computed among the two reported within-image factors.}
\label{tab:variance_partition}
\small
\begin{tabular}{lcc}
\toprule
Factor & $\eta^2$ & Relative share \\
\midrule
Direction change & 0.013 & 15.9\% \\
Location change  & 0.069 & 84.1\% \\
\bottomrule
\end{tabular}
\end{table}

Both factors explain only a small amount of the total variation, and overall variance remains dominated by image identity. This supports the main paper's interpretation that the recoverable directional component is modest relative to larger layout- and image-dependent effects.

\subsection{Signal-Injection Positive Control}
\label{app:signal_injection}

To verify that the CREG projection can recover directional signal when such signal is present, we construct synthetic relevance maps that interpolate between a position-centered signal and a deliberately conflicting directional wedge:
\begin{equation}
r_{\text{mix}} = \alpha r_{\text{dir}} + (1-\alpha) r_{\text{pos}}.
\end{equation}
Here $r_{\text{pos}}$ is centered at the target location, while $r_{\text{dir}}$ points $90^\circ$ away from the true target direction. Table~\ref{tab:signal_injection} reports representative values.

\begin{table}[t]
\centering
\caption{Signal-injection positive control. As the injected directional component increases, the compass peak shifts from following object position to following the injected direction.}
\label{tab:signal_injection}
\scriptsize
\setlength{\tabcolsep}{3pt}
\resizebox{\linewidth}{!}{%
\begin{tabular}{ccccc}
\toprule
$\alpha$ & DAE to position $\downarrow$ & DAE to injected direction $\downarrow$ & \% follows position & \% follows injected direction \\
\midrule
0.0 & 10.3 & 90.1 & 100\% & 0\% \\
0.3 & 61.0 & 40.0 & 37\% & 63\% \\
0.5 & 81.4 & 21.1 & 12\% & 88\% \\
1.0 & 91.6 & 11.1 & 0\% & 100\% \\
\bottomrule
\end{tabular}}
\end{table}

The crossover occurs around $\alpha \approx 0.3$, showing that the CREG readout can reliably recover directional structure when such structure is sufficiently present in the attribution map. This control supports the claim that weak directional results in the main paper are not caused by a failure of the projection instrument alone.

\subsection{Relational Sanity Check on Synthetic Attribution Maps}
\label{app:synthetic_sanity}

A natural concern is that DAE/EA may be biased toward target-box-only attribution, since the geometric target direction is defined from the reference object to the target object. To test this directly, we construct four synthetic relevance maps on 500 random $(A,B)$ configurations: (i) a Gaussian peak at $B$; (ii) equal Gaussian peaks at both $A$ and $B$; (iii) five Gaussian peaks along the midline segment between $A$ and $B$; and (iv) a uniform map inside $B$'s bounding box. Table~\ref{tab:synthetic_sanity} summarizes the result.

\begin{table}[t]
\centering
\small
\setlength{\tabcolsep}{5pt}
\caption{Relational sanity check on synthetic attribution maps (500 random $(A,B)$ configurations). The CREG metrics are not uniquely optimized by a target-box-only attribution form: several relation-aware synthetic patterns achieve nearly identical directional scores.}
\label{tab:synthetic_sanity}
\begin{tabular}{lcc}
\toprule
Synthetic relevance map & DAE$\downarrow$ & EA$\uparrow$ \\
\midrule
Gaussian peak at $B$ & 11.5 & $\ge .996$ \\
Equal Gaussian peaks at $A$ and $B$ & 12.3 & $\ge .996$ \\
Five Gaussian peaks along the $A$--$B$ midline & 12.5 & $\ge .996$ \\
Uniform inside $B$ bounding box & 12.3 & $\ge .996$ \\
\bottomrule
\end{tabular}
\end{table}

These results show that the CREG metrics are not uniquely optimized by a target-box-only attribution form on clean synthetic maps.

\section{Additional Cross-Model and Intervention Results}
\label{app:extra_results}

\subsection{Cross-model comparison}
\label{app:cross_model}

\begin{table*}[t]
\centering
\small
\setlength{\tabcolsep}{4.5pt}
\caption{Full cross-model comparison under the shared \creg{} readout. Bold indicates the best score for each metric within a model and dataset block. The strongest attribution plug-in is model-dependent across architectures.}
\label{tab:cross_model_appendix}
\begin{tabular}{llcccccc}
\toprule
& & \multicolumn{2}{c}{VSR} & \multicolumn{2}{c}{COCO-Pairs} & \multicolumn{2}{c}{VG-Spatial} \\
\cmidrule(lr){3-4} \cmidrule(lr){5-6} \cmidrule(lr){7-8}
Model & Method & DAE$\downarrow$ & EA$\uparrow$ & DAE$\downarrow$ & EA$\uparrow$ & DAE$\downarrow$ & EA$\uparrow$ \\
\midrule
\multirow{3}{*}{Qwen2-VL-7B}
& Trans.\ Attr.   & \textbf{61.1} & \textbf{.492} & \textbf{49.4} & \textbf{.620} & \textbf{49.2} & \textbf{.679} \\
& GradCAM         & 63.1 & .471 & 53.5 & .547 & 54.8 & .578 \\
& Attn.\ rollout  & 74.8 & .379 & 76.1 & .367 & 72.0 & .419 \\
\midrule
\multirow{3}{*}{Qwen2.5-VL-7B}
& GradCAM         & \textbf{66.8} & \textbf{.438} & \textbf{53.5} & \textbf{.533} & \textbf{53.0} & \textbf{.564} \\
& Trans.\ Attr.   & 77.9 & .329 & 70.9 & .397 & 58.6 & .547 \\
& Attn.\ rollout  & 74.9 & .388 & 75.1 & .367 & 75.0 & .389 \\
\midrule
\multirow{3}{*}{Qwen2-VL-2B}
& GradCAM         & \textbf{66.6} & \textbf{.463} & 54.3 & .537 & \textbf{51.3} & \textbf{.605} \\
& Trans.\ Attr.   & 70.2 & .421 & 68.4 & .463 & 71.7 & .483 \\
& Attn.\ rollout  & 78.2 & .346 & 81.0 & .340 & 77.7 & .365 \\
\midrule
\multirow{3}{*}{LLaVA-1.5-7B}
& GradCAM         & \textbf{66.1} & \textbf{.446} & \textbf{58.1} & \textbf{.467} & \textbf{54.8} & \textbf{.564} \\
& Trans.\ Attr.   & 74.4 & .312 & 64.6 & .457 & 59.2 & .490 \\
& Attn.\ rollout  & 73.3 & .421 & 72.5 & .387 & 62.2 & .466 \\
\bottomrule
\end{tabular}
\end{table*}

\subsection{Full intervention and cross-generational results}
\label{app:intervention_full}

Table~\ref{tab:intervention_appendix} reports the full results for the LoRA intervention and the cross-generational comparison. These results support the narrower claim made in the main paper: in our setting, accuracy gains do not reliably co-occur with lower DAE.

\begin{table*}[t]
\centering
\small
\setlength{\tabcolsep}{4.5pt}
\caption{Full results for the LoRA intervention and the cross-generational comparison. In our setting, accuracy gains do not reliably co-occur with lower DAE.}
\label{tab:intervention_appendix}
\begin{tabular}{llcccccc}
\toprule
& & \multicolumn{3}{c}{VSR} & \multicolumn{3}{c}{VG-Spatial} \\
\cmidrule(lr){3-5} \cmidrule(lr){6-8}
Setting & Model / Method & Acc & DAE$\downarrow$ & EA$\uparrow$ & Acc & DAE$\downarrow$ & EA$\uparrow$ \\
\midrule
\multicolumn{8}{l}{\textbf{LoRA training intervention (Transformer Attribution, Qwen2-VL-7B)$^\dagger$}} \\
\midrule
Baseline$^\dagger$        & --                  & .554 & 61.1 & .492 & .419 & 46.0 & .696 \\
A: Direction-supervised   & --                  & .738 & 61.7 & .492 & .561 & 51.8 & .632 \\
B: Localization           & --                  & .754 & 70.0 & .421 & .574 & 62.5 & .534 \\
C: Contrastive            & --                  & .708 & 65.2 & .458 & .611 & 48.8 & .639 \\
\midrule
\multicolumn{8}{l}{\textbf{Cross-generational comparison (each model's best-performing attribution method, 95\% CIs)}} \\
\midrule
Qwen2-VL-7B   & Chefer   & .554 & 61.1 [55, 67] & .492 & .419 & 49.2 [41, 52] & .679 \\
Qwen2.5-VL-7B & GradCAM  & .600 & 66.8 [61, 73] & .438 & .493 & 53.0 [48, 58] & .564 \\
Qwen3-VL-8B   & GradCAM  & .717 & 69.5 [63, 76] & .438 & .534 & 51.3 [47, 56] & .584 \\
\bottomrule
\end{tabular}

\vspace{0.5em}
\begin{minipage}{0.96\textwidth}
\footnotesize
$^\dagger$ The intervention-run baseline on VG-Spatial (46.0/.696) differs from the canonical run (49.2/.679) due to stochasticity in Transformer Attribution; the three intervention strategies are compared against their own run's baseline.
\end{minipage}
\end{table*}

\section{Training-Intervention Details}
\label{app:training_details}

We train all intervention models starting from Qwen2-VL-7B-Instruct with LoRA~\citep{hu2022lora}. The shared hyperparameters are:
\begin{itemize}
    \item LoRA rank $r=16$
    \item LoRA scaling $\alpha=32$
    \item Injected modules: \texttt{q\_proj} and \texttt{v\_proj}
    \item Training samples: 201 COCO-Pairs examples
    \item Epochs: 3
    \item Learning rate: $2 \times 10^{-4}$
\end{itemize}

The three strategies differ only in the auxiliary objective:

\paragraph{(A) Direction-supervised training.}
We add a compass-consistency loss that encourages the recovered compass distribution to align with the ground-truth direction:
\begin{equation}
L_{\text{compass}} = 1 - \sum_k \hat{P}(\theta_k)\cos(\theta_k - \theta_{\text{true}}),
\end{equation}
together with an entropy-based sparsity term on $\hat{P}(\theta)$.

In implementation, this loss is applied to the lightweight hidden-state compass gate used during training, not to the post-hoc multi-layer Transformer Attribution pipeline used for evaluation. Thus the direction-supervised objective uses first-order gradients through the trainable gate and LoRA parameters.

\paragraph{(B) Target-localization training.}
We encourage relevance mass to concentrate on the target object region:
\begin{equation}
L_{\text{target}} = - \log \sum_{j \in T} \mathrm{softmax}(r_j),
\end{equation}
where $T$ is the set of visual tokens whose grid-cell centers fall inside the target box.

For this objective, $r_j$ is a single-layer Grad$\times$Act surrogate computed from the target relation logit with \texttt{create\_graph=True}. We do not backpropagate through the full Chefer-style multi-layer Transformer Attribution pipeline during LoRA training.

\paragraph{(C) Relation-contrastive training.}
We add an InfoNCE-style objective over learnable relation prototypes:
\begin{equation}
L_{\text{contrastive}}
=
-
\log
\frac{\exp(\mathrm{sim}(z,p_y)/\tau)}
{\sum_c \exp(\mathrm{sim}(z,p_c)/\tau)},
\end{equation}
where $z$ is the representation of the current sample, $p_y$ is the prototype of the true relation, and $\tau$ is the temperature.

These interventions are intended only as controlled probes, not as fully optimized training recipes. The purpose of this part of the paper is therefore narrow: to test whether accuracy gains in a shared small-scale setting are reliably accompanied by better directional attribution. In our experiments, they are not.

\section{Compute and Reproducibility}
\label{app:compute}

All experiments are run on a single NVIDIA A800-80GB GPU. The complete set of experiments reported in the paper requires approximately 20 GPU-hours. All models are publicly available, and all datasets used in the paper are public. The CREG protocol is fully specified by the prompt, the projection equations, the compass hyperparameters, and the evaluation metrics given in the main paper and this appendix.


\end{document}